\documentclass[conference]{IEEEtran}
\IEEEoverridecommandlockouts
\usepackage{cite}
\usepackage{amsmath,amssymb,amsfonts}
\usepackage{algorithmic}
\usepackage{graphicx}
\usepackage{textcomp}
\usepackage{xcolor}
\usepackage{hyperref}

\def\BibTeX{{\rm B\kern-.05em{\sc i\kern-.025em b}\kern-.08em
    T\kern-.1667em\lower.7ex\hbox{E}\kern-.125emX}}
\begin{document}

\title{Drone Formation for Efficient Swarm Energy Consumption
}


\author{\IEEEauthorblockN{Shilong Guo\IEEEauthorrefmark{1},
Balsam Alkouz\IEEEauthorrefmark{2}, Babar Shahzaad\IEEEauthorrefmark{3},
Abdallah Lakhdari\IEEEauthorrefmark{4}, and
Athman Bouguettaya\IEEEauthorrefmark{5}}
\IEEEauthorblockA{School of Computer Science,
The University of Sydney\\
Australia\\
Email: \IEEEauthorrefmark{1}sguo8085@uni.sydney.edu.au,
\IEEEauthorrefmark{2}balsam.alkouz@sydney.edu.au,
\IEEEauthorrefmark{3}babar.shahzaad@sydney.edu.au,\\
\IEEEauthorrefmark{3}abdallah.lakhdari@sydney.edu.au,
\IEEEauthorrefmark{5}athman.bouguettaya@sydney.edu.au
}}

\maketitle

\begin{abstract}
We demonstrate formation flying for drone swarm services. A set of drones fly in four different swarm formations. A dataset is collected to study the effect of formation flying on energy consumption. We conduct a set of experiments to study the effect of wind on formation flying. We examine the forces the drones exert on each other when flying in a formation. We finally identify and classify the formations that conserve most energy under varying wind conditions. The collected dataset aims at providing researchers data to conduct further research in swarm-based drone service delivery. \\ Demo: \url{https://youtu.be/NnucUWhUwLs}
\end{abstract}


\section{Introduction}

Drones have received a lot of attention from the industrial and research sectors \cite{shahzaad2021top,10.1145/3460418.3479289}. Compared to terrestrial vehicles, drones are lighter, use less fuel, and are less likely to cause damage in the event of a crash \cite{jermaine2021demo}. These characteristics are assisting drones in moving beyond heavy operations, making them ripe for use in pervasive computing \cite{mueller2017drones}. A drone can provide cost-effective and reliable services to consumers and industry, ranging from photography to goods delivery \cite{9590339,lee2022autonomous}. The use of \textit{drone swarms} is a natural progression from single drones \cite{alkouz2021service}. A swarm is a group of drones that collaborate to accomplish a common goal. This new drone paradigm, i.e. swarming, is becoming available to better support people in their use of drones, and it is poised to make drones one of the key components of a pervasive computing life \cite{tahir2019swarms}.




Drone swarms are commonly used for both civil and military purposes. Examples include disaster management, environmental mapping, security, and surveillance \cite{tahir2019swarms}. A drone swarm is said to boost the single drone's capabilities for delivery applications, allowing for the simultaneous delivery of multiple packages \cite{alkouz2020formation}. Drone swarming technology is gaining popularity as evidenced by the Australian government's action plan for critical technologies, which includes drones, swarming, and collaborative robots as one of their initial focus technologies\footnote{\url{https://www.pmc.gov.au/resource-centre/domestic-policy/drones-swarming-and-collaborative-robots}}. Researchers all over the world are studying various aspects of drone swarms, such as communication, collaboration, control, and formations \cite{bacco2017uavs}. The study of \textit{drone swarm formation} for civil applications, such as delivery services, is of particular interest.


Formation flying is a natural behavior of birds \cite{cutts1994energy}. When migrating, geese fly in V formation to save energy from the lift forces generated by neighboring birds. Inspired by this natural phenomenon, military airplanes fly in a formation for a better view and protection \cite{haissig2004military}. Therefore, it is essential to investigate the formation aspect of drone swarms. For example, drone swarms can travel in a formation for better views during a search and rescue mission. Similarly, when long-distance deliveries are required, formation flying can extend the battery life of the drones, which has limited battery power (generally 30 minutes flight time) \cite{alkouz2020formation}. Furthermore, on windy days, drones in a swarm can shield each other from excessive drag forces by flying in a formation \cite{liu2022constraint}. Valuable items carried by a drone can also be protected by selecting the most advantageous position within a formation to increase robustness against wind-induced failures \cite{alkouz2022density}. Drones formations are also means to facilitate energy sharing during flights \cite{alkouz2022flight}.


We propose to demonstrate formation flying and collect a Drones Formation Dataset (DFD). This dataset is paramount to studying formation flying and its implication for a wide range of drone-based services. Existing formation flying research mainly relies on simulations to assess its impact on drone-based services \cite{alkouz2020formation}. To the best of our knowledge, there is no publicly available dataset of drone swarm formation flying and its impact on swarm energy consumption and stability. Therefore, we present the DFD dataset to advance research in swarm-based drone services.


The DFD dataset captures the dynamics of drone swarm formation flying. All drones were flown autonomously in an indoor testbed using a custom-built image-based global positioning system. We study the impact of four different flying formations on the individual drones and on the overall swarm energy consumption. We further investigate the impact of wind and the performance of each formation in reducing the impact of wind. Furthermore, we deduce the effect of different initial battery states on the full-flight energy consumption rate. Finally, we plot a sample of the collected data for analysis. This dataset could be used to determine which drone consumes the most or least energy depending on its position in the formation.


\begin{figure*}[h!]
\minipage{0.38\textwidth}
\centering
  \includegraphics[width=0.92\linewidth]{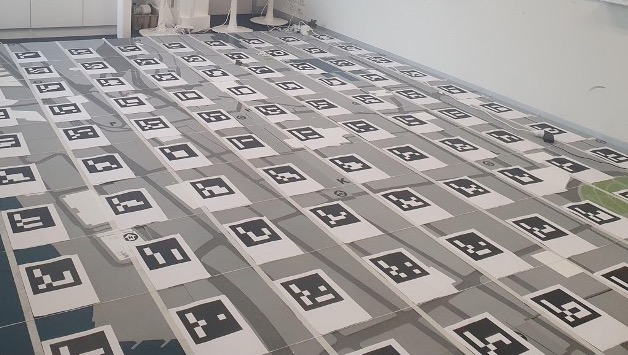}
  \caption{ArUco Markers Aided Indoor Positioning}\label{ArUco}
\endminipage\hfill
\minipage{0.62\textwidth}%
\centering
  \includegraphics[width=0.92\linewidth]{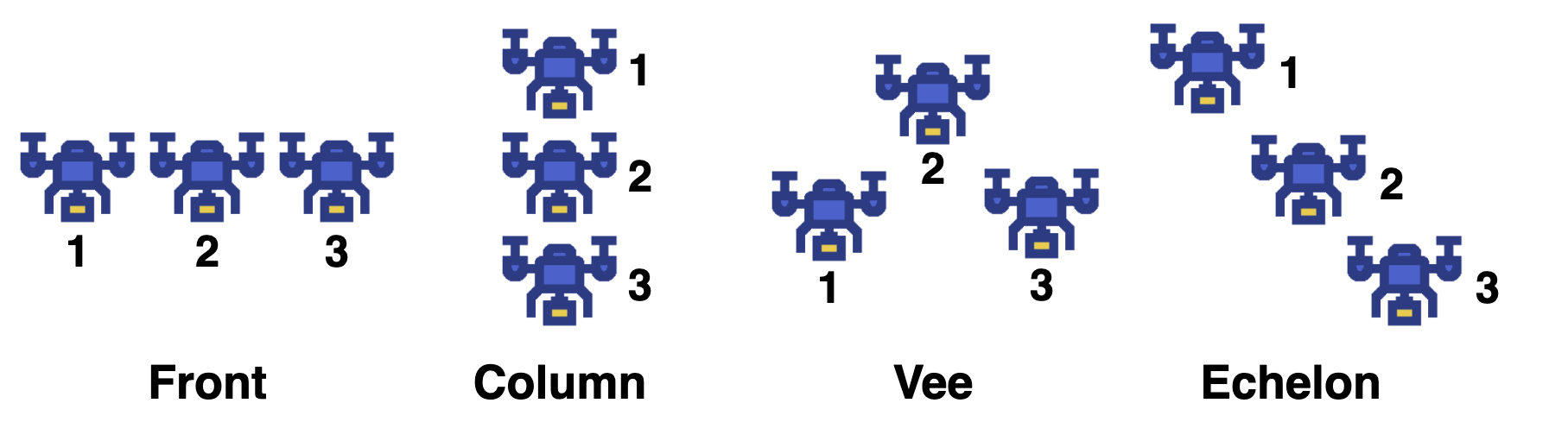}
  \caption{Different Flight Formations}\label{formations}
\endminipage\hfill
\end{figure*}
\section{Trajectory Tracking and Data Collection}
\subsection{Hardware and Software Setup}
\subsubsection{Hardware and Data Collection Environment.}
Outdoor testing poses significant safety risks, especially when dealing with swarms of drones. Therefore, we demonstrate and collect the dataset in an indoor drone testbed environment. In indoor testbeds, smaller drones are usually used for experiments. In our indoor testbed, we use the DJI Edu Tello drones\footnote{\url{https://www.ryzerobotics.com/tello-edu}} that are equipped with swarming capabilities for the experiments. 
DJI Tello drones are equipped with two cameras: (1) a front camera and (2) a bottom-facing infrared camera. The drone's SDK is originally programmed to locate itself with mission pads that come with the drone. However, this positioning system suffers from precision and tracking errors, especially with swarms. Therefore, we create our own indoor positioning system using ArUco markers. ArUco markers are similar to QR codes that allow algorithms to decode the information stored in the marker quickly. Fig. \ref{ArUco} illustrates the indoor testbed with 120 ArUco markers, 10 on the x-axis and 12 on the y-axis, with a 40 cm distance from each other. We use the drone's bottom-facing infrared camera stream to read the markers as described in the software setup below. We also use fans and an anemometer to simulate wind in the indoor testbed. \looseness=-1


\subsubsection{Software and Indoor Positioning System Configuration.}
When flying in a swarm, a router or access point is needed to communicate with all of the drones. Each Tello drone receives the swarm commands using a Python script that the router sends to it. Each drone in the swarm is first individually reconfigured and switched to station mode. We connect to Tello's WiFi using the Packet Sender\footnote{https://packetsender.com/} software and switch the drone into SDK mode for sending swarming commands. All drones in the swarm send back their video stream using User Datagram Protocol forwarding. Then, the code identifies the ArUco marker ID where a drone is currently hovering and its distance from that marker in each direction using OpenCV. Since each marker corresponds to a global coordinate, the system adds/subtracts the drone's distance from a marker to calculate its precise coordinate. Once we have the drone's global coordinate, we compare it to the drone's required coordinate to navigate the drone to its desired position. An adjustment function alters the drone's x and y velocity to get it on track to the desired position. 



\subsection{Collected Data}
We collect a dataset for four formations, namely Column, Front, Vee, and Echelon (Fig. \ref{formations}). For each formation, we fly a swarm of three drones for one minute to capture the effect of formation on different positions within a formation. The swarms are flown under three wind conditions (no wind, front wind, and side wind). For wind experiments, the wind speed is set to 6.1 km/h. The attributes collected include the degree of the attitude (pitch, roll, yaw), speed of axes (x,y, and z), lowest and highest temperatures in degree Celsius, battery percentage, barometer measurement, amount of time the motor has been used, and acceleration in axes (x,y, and z). 

\section{Data Analysis and Evaluation}
\begin{figure}[h!]
    \centering
    \includegraphics[width=0.9\linewidth]{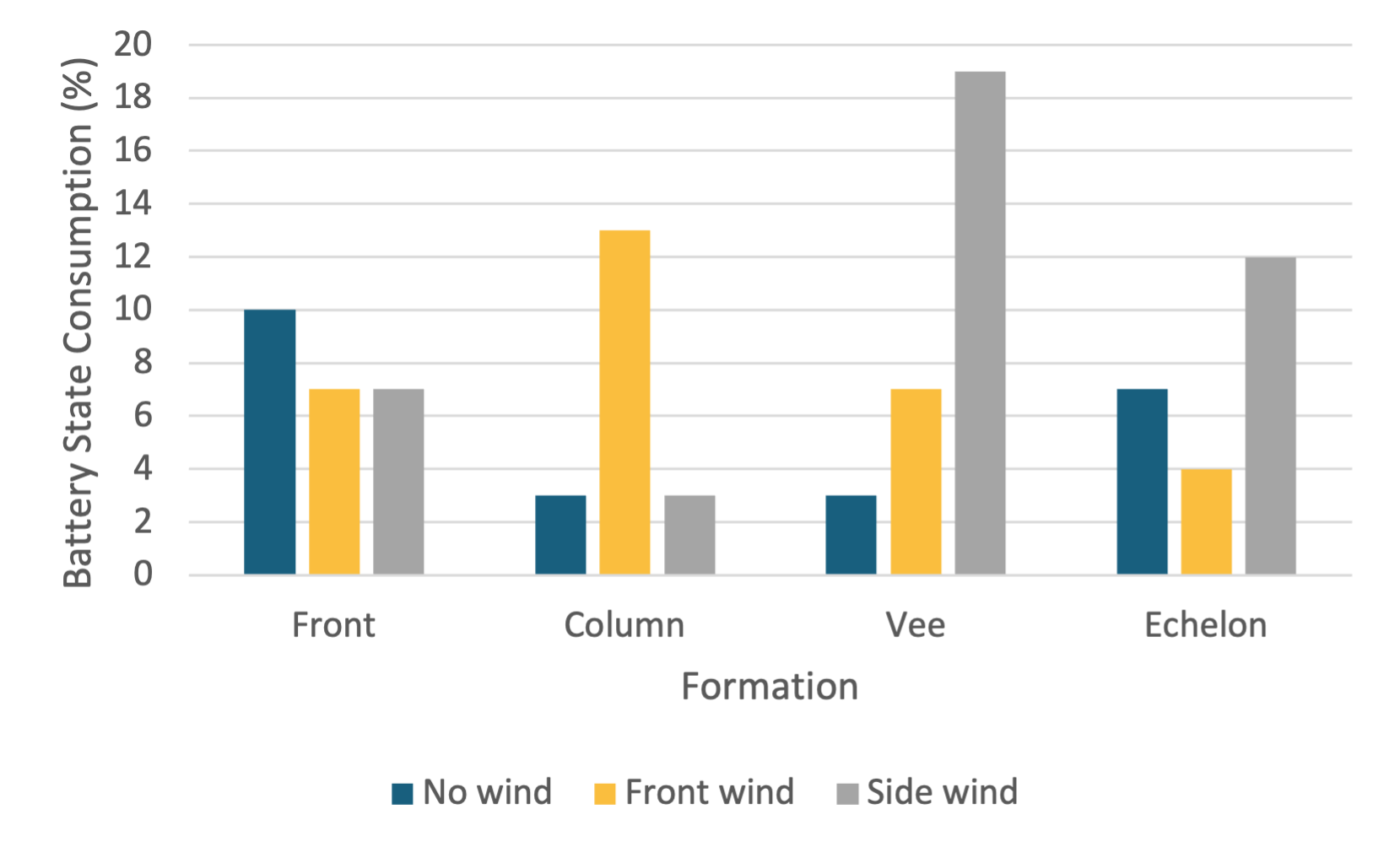}
    \vspace{-0.4cm}
    \caption{Battery Consumption of Different Formations}
    \label{formationsbattery}
\end{figure}
We present a sample analysis of the collected dataset in terms of battery consumption and swarm acceleration under different formations and wind conditions. Fig. \ref{formationsbattery} represents the average energy consumption of all drones in different formations and wind conditions during a one-minute flight. This figure enables researchers to choose the best formation if the goal is to optimize the entire swarm rather than an individual drone within a formation. As shown in the figure, the Vee formation consumes the least amount of energy without wind due to generated upwash forces helping neighboring drones \cite{mirzaeinia2019energy}. This behavior aligns with the behavior observed by birds and reported in \cite{cutts1994energy}. However, the Vee formation consumes the most energy with a side wind. Similarly, the Column formation consumes the least energy with side winds. This decrease in energy consumption is due to side winds providing lift forces to all drones as they are exposed to the wind.





\begin{figure}
    \centering
    \includegraphics[width=0.75\linewidth]{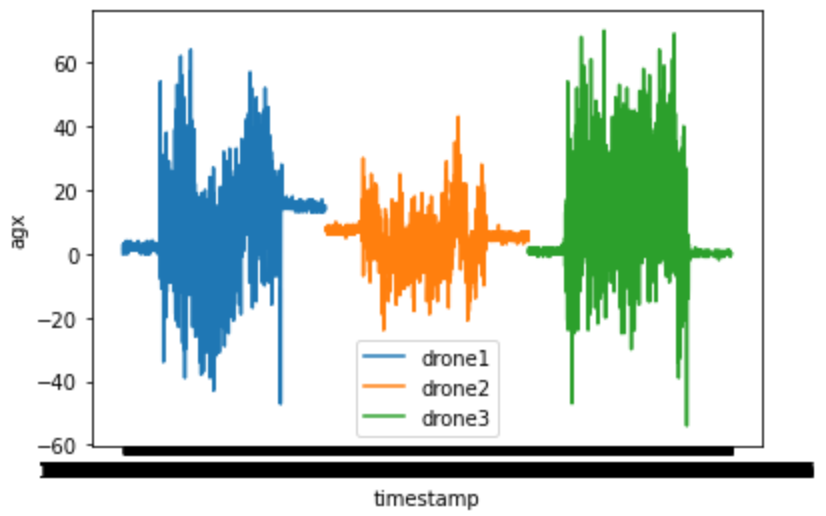}
    \caption{Front Formation X-axis Acceleration (No Wind)}
    \label{accelrationnowind}
\end{figure}

\begin{figure}
    \centering
    \includegraphics[width=0.75\linewidth]{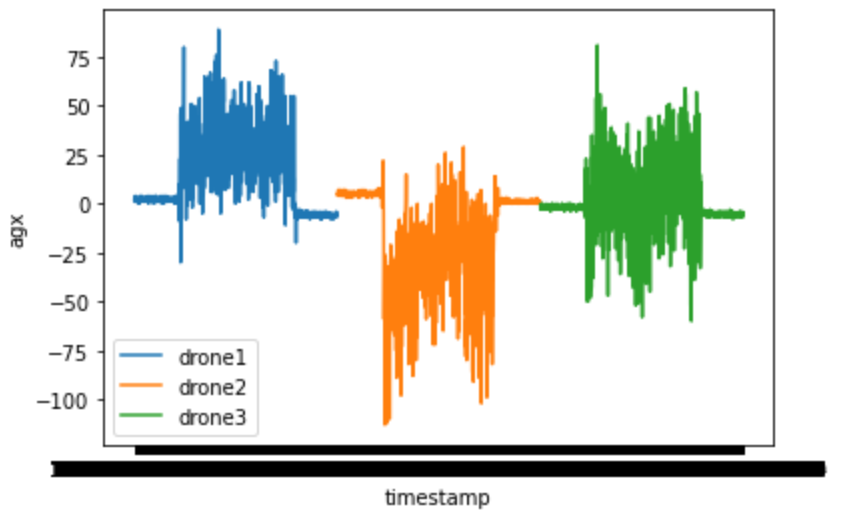}
    \caption{Front Formation X-axis Acceleration (Front Wind) }
    \label{accelrationwind}
\end{figure}

\begin{figure}
    \centering
    \includegraphics[width=0.65\linewidth]{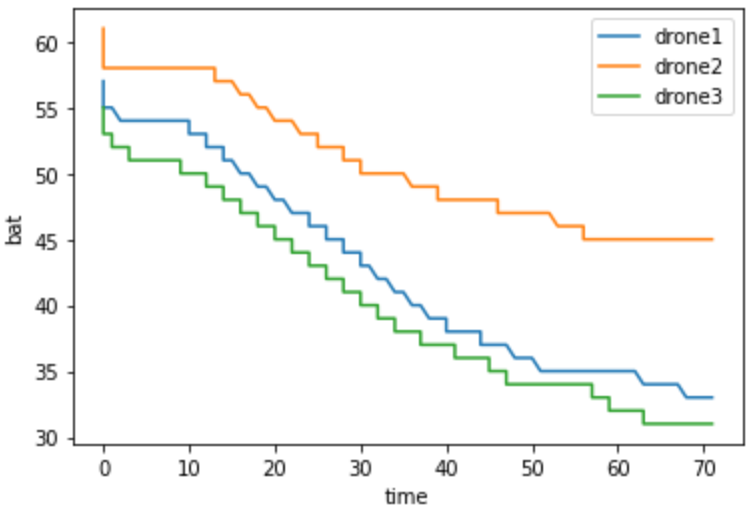}
    \caption{Battery State (Echelon, Side Wind)}
    \label{battery}
\end{figure}

\begin{figure}
    \centering
    \includegraphics[width=0.72\linewidth]{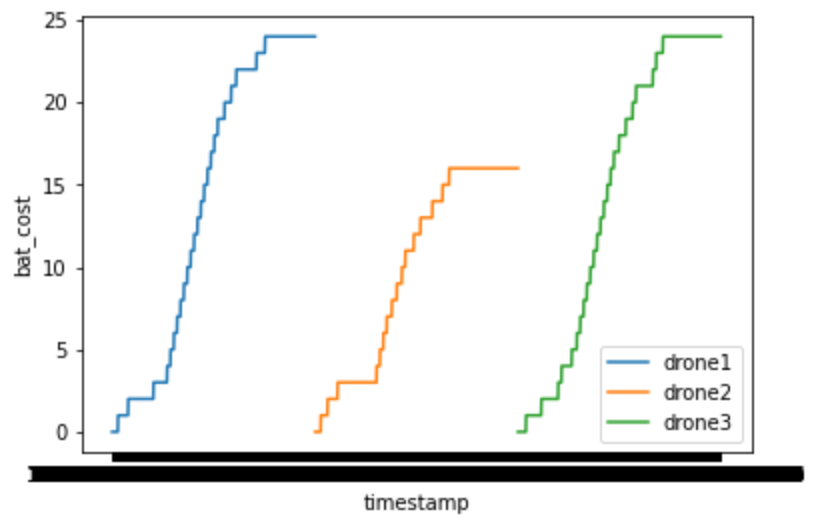}
    \caption{Battery Cost (Echelon, Side Wind)}
    \label{batterycost}
\end{figure}

We analyze some formations to study their effect on individual drones within a swarm. Figs \ref{accelrationnowind} and \ref{accelrationwind}  represent the acceleration in the x-axis for the drones in a Front formation under no wind and front wind. As shown in Fig. \ref{accelrationnowind},  the middle drone produces downwash forces affecting the acceleration and stability of drones 1 and 3. With front wind, as represented in Fig.\ref{accelrationwind}, the acceleration scale, as shown on the graph's y-axis, gets larger, indicating higher wind resistance. Figs \ref{battery} and \ref{batterycost} represent the battery level and cost of the three drones in an Echelon formation under the side conditions. As shown in the graph, although drone 3 is expected to consume less energy as it is farther away from the wind source and is partially hidden by drones 1 and 2, it consumes the most energy, which could be due to the initial battery state of the drone leading to higher consumption. This behavior reflects that drones with a full charge or higher charge capacity consume energy less than drained battery drones.

\section*{Acknowledgment}
This research was partly made possible by LE220100078 grant from the Australian Research Council. The statements made herein are solely the responsibility of the authors.

\bibliographystyle{IEEEtran}  
\bibliography{references}


\end{document}